\documentclass[times, review, 10pt]{elsarticle}

\usepackage{amssymb}
\usepackage{multirow,amsmath,booktabs,diagbox}
\usepackage{amsmath}
\usepackage{soul}
\usepackage{threeparttable}

\usepackage{algorithm}
\usepackage{algorithmic}
\usepackage{dsfont}
\usepackage{hyperref}
\usepackage{xcolor}
\newcommand{\highlight}[1]{\textcolor{black}{#1}}

\newcommand{\etal}{\textit{et al.}}
\journal{Pattern Recognition}

\begin{document}

\begin{frontmatter}



\title{CurvNet: Latent Contour Representation and Iterative Data Engine for Curvature Angle Estimation}



\author[1,2,3]{Zhiwen~Shao}
\ead{zhiwen_shao@cumt.edu.cn}
\author[1,2]{Yichen Yuan}
\ead{yuanyichen@cumt.edu.cn}
\author[3]{Lizhuang~Ma}
\ead{ma-lz@cs.sjtu.edu.cn}
\author[4,5]{Xiaojia~Zhu\corref{correspondingauthor}}\cortext[correspondingauthor]{Corresponding author.}
\ead{xiaojia_zhu@foxmail.com}

\address[1]{School of Computer Science and Technology, China University of Mining and Technology, Xuzhou 221116, China}
\address[2]{Mine Digitization Engineering Research Center of the Ministry of Education, Xuzhou 221116, China}
\address[3]{School of Computer Science, Shanghai Jiao Tong University, Shanghai 200240, China}
\address[4]{Xuzhou Central Hospital/The Xuzhou Clinical School of Xuzhou Medical University, Xuzhou 221009, China}
\address[5]{Xuzhou Rehabilitation Hospital/The Affiliated Xuzhou Rehabilitation Hospital of Xuzhou Medical University, Xuzhou 221003, China}


\begin{abstract}
Curvature angle is a quantitative measurement of a curve, in which Cobb angle is customized for spinal curvature. Automatic Cobb angle measurement from X-ray images is crucial for scoliosis screening and diagnosis. However, most existing regression-based and segmentation-based methods struggle with inaccurate spine representations or mask connectivity and fragmentation issues. Besides, landmark-based methods suffer from insufficient training data and annotations. To address these challenges, we propose a novel curvature angle estimation framework named CurvNet including latent contour representation based contour detection and iterative data engine based image self-generation. Specifically, we propose a parameterized spine contour representation in latent space, which enables eigen-spine decomposition and spine contour reconstruction. Latent contour coefficient regression is combined with anchor box classification to solve inaccurate predictions and mask connectivity issues. Moreover, we develop a data engine with image self-generation, automatic annotation, and automatic selection in an iterative manner. By our data engine, we generate a clean dataset named Spinal-AI2024 without privacy leaks, which is the largest released scoliosis X-ray dataset to our knowledge. Extensive experiments on public AASCE2019, our private Spinal2023, and our generated Spinal-AI2024 datasets demonstrate that our method achieves state-of-the-art Cobb angle estimation performance. Our code and Spinal-AI2024 dataset are available at \textit{https://github.com/Ernestchenchen/CurvNet} and \textit{https://github.com/Ernestchenchen/Spinal-AI2024}, respectively. 
\end{abstract}










\end{frontmatter}



\section{Introduction}

\begin{figure}[!b]
\centering\includegraphics[width=0.8\linewidth]{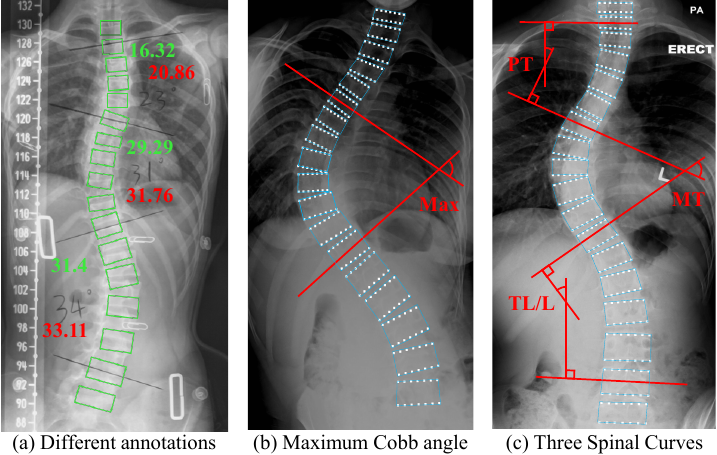}
\caption{Illustration of Cobb angle measurement. There are three categories of scoliosis: proximal thoracic (PT) curve, main thoracic (MT) curve, and thoracolumbar/lumbar (TL/L) curve, as shown in (c). In (a), handwritten 23°, 31°, and 34° are manually measured by experts, angles in green are measured based on rectangular boxes (also in green), and angles in red are measured based on our proposed contour boxes composed by spinal landmarks (drawn in (b)). (b) and (c) show two types of evaluations: maximum angle of three spinal curves, three regional angles, respectively.}
\label{page1fig1}
\end{figure}

Automatic measurement of curvature angle is important for curved objects. Due to critical role of spine in human body, Cobb angle estimation of scoliosis, a common spinal deformity disease in human beings, has gained increasing attention in the field of artificial intelligence and computer vision~\cite{zhu2025mgscoliosis}. %
Early detection and treatment of scoliosis can help reduce the need for surgical interventions \cite{prabhu2012automatic}. %
The Cobb angle is one of the most widely used clinical measures for evaluating and diagnosing the severity of scoliosis. It is defined as the angle between parallel lines of the top of the upper end vertebra (the most tilted vertebrae on the curve) and the bottom of the lower one~\cite{mcleod2022anesthesia}. There are three regional spinal deformities associated with different spinal segments: proximal thoracic (PT), main thoracic (MT), and thoracolumbar/lumbar (TL/L), as illustrated in Fig.~\ref{page1fig1}(c).

Currently, manual Cobb angle measurement on radiographic images like X-ray images by experts is still the mainstream way in medical institutions. However, it is time-consuming and often has potential errors, 
since it can be affected by spinal selection and inter-observer variability, which can be observed from the distances between manual angles and angles by landmarks in Fig.~\ref{page1fig1}(a).   
It has been validated that manual annotation of the Cobb angle exhibits fluctuations ranging from 2.8 to 8 degrees~\cite{sun2022comparison}. Therefore, developing a reliable automatic Cobb angle measurement method is crucial.

Traditional automatic
Cobb angle measurement methods rely on hand-crafted features~\cite{wu2017automatic}, limiting model performance and generalization. In recent years, some regression-based methods use  rectangular boxes as the spine representation~\cite{khanal2020automatic}, which suffer from reduced accuracy when spinal blocks are wedge-shaped or irregular, as shown in Fig.~\ref{page1fig1}(a). 
Segmentation-based methods are sensitive to image quality~\cite{lin2020seg4reg,wu2023automated}, tending to predict connected or damaged spinal masks for low-quality inputs. Landmark-based methods necessitate vast amounts of training data and annotations~\cite{xu2021graph,yi2020vertebra},
which are costly and time-consuming to collect and label, and are exacerbated by data scarcity due to patient privacy concerns.

In this paper, we distill the above limitations as three main issues, and tackle them from three aspects: (a) How to achieve more accurate spine representation? (b) How to generate sufficient scoliosis images while protecting patient privacy? (c) How to obtain accurate annotations at low costs?

\textit{Parameterized Spine Representation.}
We propose a novel automatic Cobb angle measurement framework named CurvNet. We combine regression-based and landmark-based techniques, by using a set of spinal landmarks to represent the bounding box of spinal segments. Inspired by low-rank approximation parameterized contour representation in object detection~\cite{su2024lranet}, 
we use latent space to learn eigen-spines from spine contours of training data, and reconstruct spine contours via latent contour coefficients. 
To suppress inaccurate predictions and mask connectivity, we integrate latent contour coefficient regression with anchor box classification to obtain spinal segment contours, in which sparse and dense assignments are integrated to more precisely detect spinal segments.

\textit{Privacy-Preserving Scoliosis Generation.} We explore 
the latent diffusion model (LDM)~\cite{rombach2022high} for scoliosis image self-generation. Specifically, we train LDM on our private dataset Spinal2023, utilizing a diffusion process to map raw images from pixel to latent space. During inference, the trained LDM with text-conditioned prompts can generate X-ray images. To mitigate privacy leaks~\cite{carlini2023extracting}, we employ data augmentation during LDM training, promote data balance, and introduce a privacy review~\cite{packhauser2023generation} to exclude potentially privacy-compromising samples based on structural and pixel similarities, coupled with manual verification.

\textit{Cost-Effective Semi-Supervised Annotation.} 
Inspired from semi-supervised pseudo-labeling~\cite{rizve2021defense} and multi-stage interactive annotation~\cite{kirillov2023segment}, we propose a new data engine, which drives an iterative pipeline with automatic annotation and selection.  
It consists of four stages: pseudo-labeling, auto-annotation, manual-assisted annotation, and privacy review.
At each iteration, spine contour detection network is first re-trained using current dataset, and its trained model is used for pseudo-labeling so as to further update the dataset. This iterative pipeline culminates in our open-source clean dataset named Spinal-AI2024, which is the largest released scoliosis X-ray dataset to our knowledge.

The main contributions of this paper are threefold:

\begin{itemize}
\item We propose a novel low-rank approximation parameterized spine representation for curvature angle estimation, in which latent contour coefficient regression and anchor box classification with sparse and dense assignments are combined to detect irregular contours of spinal segments.

\item We propose a new data engine with semi-supervised labeling and privacy review in an iterative manner, which enables the generation of large-scale clean scoliosis X-ray images without privacy leaks. To our knowledge, our open-source Spinal-AI2024 is the largest released scoliosis X-ray dataset.

\item Extensive experiments on AASCE2019, Spinal2023, and Spinal-AI2024 datasets show that our method outperforms state-of-the-art automatic Cobb angle measurement works for both maximum angle and three regional angles.
\end{itemize}

\section{Related Work}
\label{sec:Related Work}

\subsection{Scoliosis Cobb Angle Measurement}

Earlier methods~\cite{sun2017direct,wu2017automatic} adopt traditional machine learning to approximate the relationship between image features and Cobb angles. However, these methods suffer from performance and generalization issues due to loss of high-resolution details through downsampling~\cite{yi2020vertebra} or reliance on hand-crafted features. Recently, deep learning based methods~\cite{shao2025mol} have gained increasing attention. 

Regression-based methods~\cite{khanal2020automatic,targ2016resnet,wu2018automated,wang2019accurate} aim to learn a direct mapping between the spinal shape representation and the Cobb angle, without explicit geometric calculations. However, these approaches have limited performance for wedge-shaped or irregular vertebrae due to rectangular shape assumption. 

    Segmentation-based methods~\cite{lin2020seg4reg,chen2022automating} usually rely on U-Net~\cite{ronneberger2015u} or its modifications~\cite{horng2019cobb,wu2023automated}, 
    by performing pixel-level classification and then combining pixels as spinal regions. 
    However, these methods are sensitive to image quality, and often require pre-processing to address noise, resolution, and contrast issues~\cite{chen2022automating}.

    Landmark-based methods~\cite{sun2019deep,xu2021graph,zhou2023vertebral,yi2020vertebra,guo2022cobb} focus on localizing spinal landmarks, which often requires vast training data. In this paper, we overcome the data scarcity issue by predicting landmarks through latent contour coefficient regression, leveraging our proposed data engine with automatic annotation and selection. 

\subsection{Medical Image Generation}

Deep learning relies heavily on numerous labeled training data. 
In the medical image field, there is often a scarcity of annotated data due to cumbersome, time-consuming, and expensive image acquisition and annotation. 
    To alleviate this limitation, previous works use data perturbation or data synthesis. 

    Generative models
    like generative  adversarial networks (GANs) \cite{goodfellow2020generative} have demonstrated the capability to produce realistic, high-resolution images. 
    Recent studies \cite{carlini2023extracting} indicate that diffusion models \cite{rombach2022high} possess higher generation capacity than GANs. 
    Packh{\"a}user \etal~\cite {packhauser2023generation} utilized latent diffusion model (LDM) to generate chest X-ray images, and proposed a sampling strategy that safeguards the privacy of sensitive biometric information. However, sharing sensitive data under stringent privacy regulations remains a critical challenge in medical research \cite{esteva2019guide}. 
    In this paper, we also leverage LDM to generate  scoliosis X-ray images. Additionally, we propose a privacy-auditing module combined with pseudo-label training, utilizing a comprehensive privacy similarity mechanism to filter out images that may expose patient privacy.

\subsection{Semi-Supervised Labeling}
 Semi-supervised labeling methods can be categorized into consistency regularization and pseudo-labeling approaches. Consistency regularization methods \cite{honari2018improving} rely on modality-specific augmentation for regularization. Pseudo-labeling methods \cite{li2021synthetic} predict labels for unlabeled data and train the model in a supervised manner. 

    Shi \etal~\cite{shi2018transductive} 
    adopted local neighborhood density-based confidence scores to facilitate pseudo-labeling. Inspired by noise correction \cite{yi2019probabilistic}, Wang \etal~\cite{wang2020repetitive} updated pseudo-labels by an optimization framework. Xie \etal~\cite{xie2020self} demonstrated the effectiveness of self-training in supervised classification tasks. 
    However, these methods struggle with flawed pseudo-labels.
    Recently, Rizve \etal~\cite{rizve2021defense} selected high-quality pseudo-labels from noisy samples based on uncertainty. 
    Kirillov \etal~\cite{kirillov2023segment} built a data engine to augment data through manual-assisted, semi-automatic, and fully automatic phases. In this paper, we employ a similar semi-automatic annotation pipeline to expand datasets, in which pseudo-label selection and manual assistance are integrated for quality improvement.
        
\begin{figure}[!t]
\centering
\includegraphics[width=1.0\linewidth]{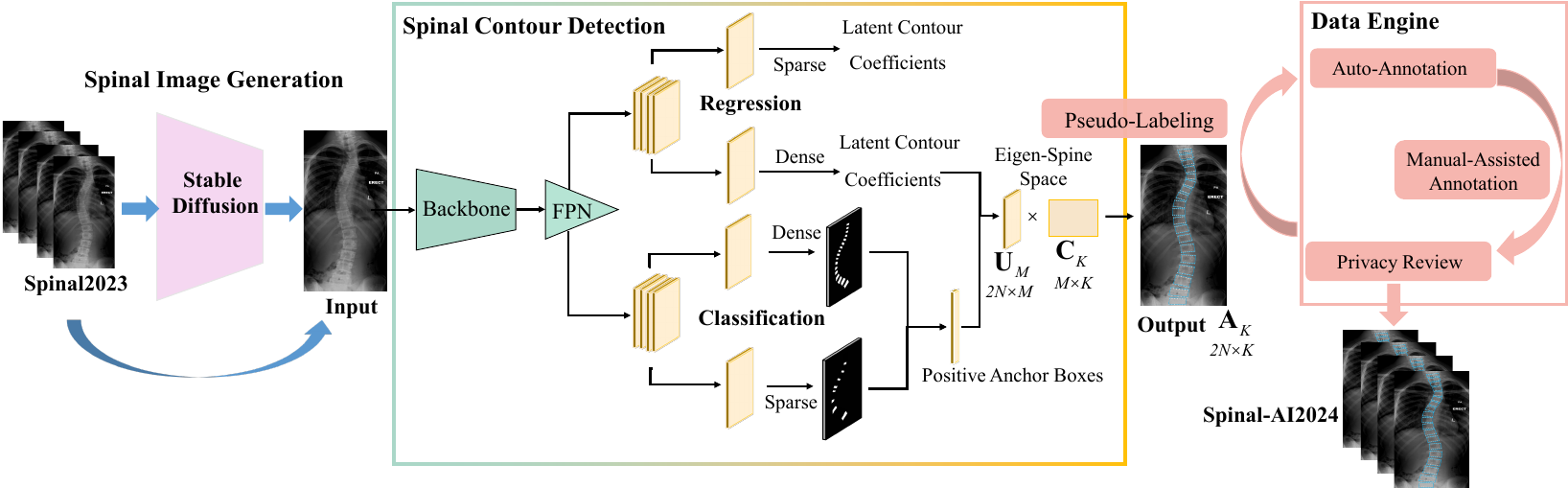}
\caption{The overall structure of our CurvNet framework, which consists of spinal image generation, spinal contour detection, and data engine. Our private Spinal2023 dataset is used for training 
spinal image generation model, and also used for training initial model of spinal contour detection
network. During each round of data engine, 
trained spinal contour detection network is adopted for pseudo-labeling, and selected data with annotations are backward used to fine-tune the spinal contour detection network. By employing the data engine to select and annotate generated images, we obtain the Spinal-AI2024 dataset.
}
\label{page3fig3}
\end{figure}

\section{Methodology}

The architecture of our framework is illustrated in Fig.~\ref{page3fig3}. 
Given our private Spinal2023 dataset, we first train Stable Diffusion~\cite{rombach2022high} to generate vast spinal X-ray images. Specifically, 
the generation starts from the original pixel space, undergoing diffusion. The textual condition and pixel features are then fed into the latent space, where cross-attention modules produce the final output.

Also, we use Spinal2023 to train an initial model of spinal contour detection network. This network adopts ResNet50 \cite{he2016deep} as backbone, and uses feature pyramid network (FPN) \cite{lin2017feature} to extract multi-scale features. Then, it uses regression and classification branches to predict latent contour representation
and positive anchor boxes, respectively. In both branches, sparse and dense assignments are combined to improve the detection performance. Finally, we select latent contour coefficients
corresponding to positive anchor boxes to reconstruct spinal segment contours.

The data engine integrates a loop with a screening mechanism. The screening process encompasses pixel-level privacy checks, sample-level legality reviews, and segment-level quality assessments. At each iteration, spine contour detection network is fine-tuned using
current data, and its trained model is further used for pseudo-labeling so as to update the data and annotations.

\subsection{Spinal Contour Detection via Latent Contour Representation}
\subsubsection{Parameterized Spine Representation}
To address inaccuracy of rectangular box based spine representation in prior works, we utilize contour
box composed by spinal landmarks to represent spinal segments. Given that spinal segments generally exhibit similar structural properties and notable correlations among their contours, we develop a new parameterized spinal contour representation method based on such correlations from training data.

In particular, the boundary of a spinal segment is composed of $N$ vertices, i.e. landmarks, which can be represented as a vector $\mathbf{a} = [ x_1,y_1, \cdots , x_N , y_N  ]\in \mathbb{R}^{2N\times 1}$ containing x and y coordinates. Then, we construct a spinal contour matrix 
$\mathbf{A} = [\mathbf{a}_1,\mathbf{a}_2,\cdots,\mathbf{a}_L ] 
 \in \mathbb{R}^{2N \times L}$ from training data containing $L$ spinal segment instances. 
  Subsequently, we employ singular value decomposition (SVD) to extract the underlying structural relationship among $\mathbf{A}$:
\begin{equation}  
\mathbf{A = U\Sigma V}^T= \sigma_1 \mathbf{u}_1 \mathbf{v}_1^T + \cdots + \sigma_r \mathbf{u}_r
\mathbf{v}_r^T,
\label{A:decompose}
\end{equation}  
where $\mathbf{U} =  [\mathbf{u}_1,\mathbf{u}_2,\cdots,\mathbf{u}_{2N}]\in \mathbb{R}^{2N \times 2N}$and $\mathbf{V} = [\mathbf{v}_1,\mathbf{v}_2,\cdots,\mathbf{v}_L]\in \mathbb{R}^{L \times L}$ are orthogonal matrices. $\Sigma \in \mathbb{R}^{2N \times L}$
is a diagonal matrix composed of singular values $\sigma_1 \geq \sigma_2 \geq \cdots \geq \sigma_r > 0$, in which $r$ is the rank of $\mathbf{A}$.

Each singular vector represents an orthogonal direction in the eigen-space. By retaining the first $M$ singular values and their corresponding vectors while truncating the remaining low-rank components in Eq.~\eqref{A:decompose}, we can approximate $\mathbf{A}$ as 
\begin{equation}
    \mathbf{A}_M = \sigma_1 \mathbf{u}_1 \mathbf{v}_1^T + \cdots + \sigma_M \mathbf{u}_M \mathbf{v}_M^T=\mathbf{U}_M \mathbf{\Sigma}_M \mathbf{V}_M^T,
\end{equation}
where $\mathbf{A}_M$ is the best rank-M approximation of $\mathbf{A}$, since it minimizes the Frobenius norm $\|\mathbf{A} - \mathbf{A}_M\|_F$
\cite{blum2020foundations}.

Then, we define $\mathbf{C}_M = \mathbf{\Sigma}_M \mathbf{V}_M^T=[
\mathbf{c}_1 , \mathbf{c}_2 , \cdots , \mathbf{c}_L ] 
 \in \mathbb{R}^{M \times L}$, such that $\mathbf{A}_M$ can be represented as
\begin{equation}
\mathbf{A}_M = [ \mathbf{U}_M\mathbf{c}_1 , \cdots , \mathbf{U}_M\mathbf{c}_L]=[ \tilde{\mathbf{a}}_1 , \cdots , \tilde{\mathbf{a}}_L],  
\label{A:approximated}
\end{equation}
where $\tilde{\mathbf{a}}_i = \mathbf{U}_M\mathbf{c}_i = [ \mathbf{u}_1 , \mathbf{u}_2 , \cdots , \mathbf{u}_M ] \mathbf{c}_i$ is the approximation of $\mathbf{a}_i$.

We refer to $\mathbf{u}_1, \ldots, \mathbf{u}_M$ as \textit{eigen-spines}, since they are eigen-vectors of the matrix $\mathbf{A A^T}$ and can be viewed as principal spinal segment shapes.

For any $2N$-dimensional spinal segment contour $\mathbf{a}$, we can project it onto the eigen-spine space: 
\begin{equation}
 \mathbf{c} =  \mathbf{U}_M^T \mathbf{a},
\end{equation}
where \highlight{the eigen-spine matrix $\mathbf{U}_M \in \mathbb{R}^{2N \times M}$ projects $\mathbf{a}\in \mathbb{R}^{2N\times 1}$ into the latent space, and} $\mathbf{c}\in \mathbb{R}^{M\times 1}$ is the latent contour coefficient vector. 


\subsubsection{Spinal Contour Detection Network}

As shown in Fig.~\ref{page3fig3}, our spinal contour detection network contains regression and classification branch, in which the overall loss is defined as
\begin{equation}
\mathcal{L} = \lambda_{reg} \mathcal{L}_{reg}+\lambda_{cls} \mathcal{L}_{cls},
\label{eq:lra_total}
\end{equation}
where $\mathcal{L}_{reg}$ and $\mathcal{L}_{cls}$ are the losses for the regression and classification branches, respectively, and $\lambda_{reg}$ and $\lambda_{cls}$ control their importances.
$\mathcal{L}_{cls}$ 
is composed by the spine region loss $\mathcal{L}_{sr}$ in dense assignment and the sparse sampling region loss $\mathcal{L}_{ssr}$ in sparse assignment:
\begin{equation}
\mathcal{L}_{cls} = \mathcal{L}_{sr} + \mathcal{L}_{ssr},
\end{equation}
in which $\mathcal{L}_{sr}$ and $\mathcal{L}_{ssr}$ use the form of cross-entropy loss and focal loss,
respectively.
$\mathcal{L}_{reg}$ is defined as 
\begin{equation}
\label{eq:L_reg}
\mathcal{L}_{reg} = \sum_i \mathds{1}[i\in ps] l_1 (\tilde{\mathbf{a}}_i, \mathbf{a}_i),
\end{equation}
where $l_1$ denotes smooth-L1 loss, $ps$ denotes the positive sample region obtained by incorporating sparse and dense assignments, and $\mathds{1}$ denotes indicator function, outputting 1 when the condition is valid (the $i$-th point is within $ps$ in Eq.~\eqref{eq:L_reg}) and outputting 0 otherwise. After combining the outputs of regression and classification branches, we can obtain a latent contour coefficient matrix $\mathbf{C}_K\in \mathbb{R}^{M\times K}$. The predicted spinal segment locations are calculated as
\begin{equation}
    \mathbf{A}_K = \mathbf{U}_M \mathbf{C}_K,
\end{equation}
where $K$ denotes the number of detected spinal segments.

\begin{algorithm}[!t]
\caption{The overall procedure of our iterative data engine. }
\label{alg:procedure}
\begin{algorithmic}[1] 
\REQUIRE Labeled dataset $D_{spinal2023}$, unlabeled dataset $D_{selected}$ generated by Stable Diffusion, sample selection index set $V_{selected}$ initialized as empty set, and spinal contour detection network with parameters $\Theta$.
\ENSURE 
Labeled dataset $D_{spinalAI2024}$.
\STATE Train spinal contour detection network on $D_{spinal2023}$, and obtain initial parameters $\Theta^0$.

\STATE Set $i \gets 1$.  
\label{alg:procedure:2}
\WHILE{$V_{selected}$ is not converged}
\label{alg:procedure:3}
\STATE Pseudo-label samples in $D_{selected}$ using $\Theta^{i-1}$ with $\tau_c$-based positive spinal segment instance selection.
\STATE 
Update $V_{selected}$ by auto-annotation, manual-assisted annotation, and privacy review.
\STATE $D_{spinalAI2024} \gets \{s^j|s^j\in D_{selected}, v^j=1 \}$.
\STATE Fine-tune network from $\Theta^{i-1}$ using samples from $D_{spinal2023} \cup D_{spinalAI2024}$, and obtain new parameters $\Theta^{i}$.

\STATE Set $i \gets i+1$.
\ENDWHILE
\label{alg:procedure:9}
\STATE Return 
$D_{spinalAI2024}$.
\\

\end{algorithmic}
\end{algorithm}

\subsection{Iterative Data Engine with Image Self-Generation}

Our data engine is composed of iteratively four stages: pseudo-labeling, auto-annotation, manual-assisted annotation, and privacy review. 
Before starting the data engine, we use Stable Diffusion~\cite{rombach2022high} to generate an initial set of spinal X-ray images, 
as shown in 
Algorithm~\ref{alg:procedure}. \highlight{We use an early stopping strategy for the iterative process from Step~\ref{alg:procedure:3} to Step~\ref{alg:procedure:9}, in which the iteration stops when the number of newly selected samples changes by less than 1\% over two consecutive rounds, or after six iterations.}

\subsubsection{Pseudo-Labeling}
At the $i$-th iteration, to enable pseudo-labeling, we first train our spinal contour detection network on currently available labeled data composed of $D_{spinal2023}$ and $D_{spinalAI2024}$, in which the trained model parameters are denoted as $\Theta^{i-1}$. Then, we use $\Theta^{i-1}$ to label samples in $D_{selected}$. For the $j$-th sample $s^j$, the network predicts the locations of $K$ spinal segments $[
\mathbf{a}_1^j , \mathbf{a}_2^j , \cdots , \mathbf{a}_K^j ]$. To filter inaccurate predictions, we use a confidence threshold $\tau_c$ to select positive spinal segment instances for $s^j$:     
\begin{equation} 
\mathcal{A}^j=\{\mathbf{a}_k^j |p^j_k \geq \tau_c \},
\label{eq:confidence}
\end{equation} 
where $p^j_k$ denotes the probability/confidence of the $k$-th spinal segment output by spinal contour detection network.

\subsubsection{Auto-Annotation}
We perform segment-level and sample-level automatic selection. 
We remove spinal segment instances with area smaller than 
200 pixel$^2$, remove instances with illegal landmark coordinates like negative values or positions beyond image boundaries, and remove instances whose landmark coordinates cannot form a valid contour.
At the sample-level, we discard samples with fewer than 10 instances, in which the average number of spinal segments per image is 17 in real datasets. 
We also discard samples with spinal center distances larger than three times of the average distances. This sample-level selection can filter out samples with poor generation qualities. 
The index $v^j$ of sample $s^j$ is set to 1 in $V_{selected}$ if it is selected, and is set to 0 otherwise.

\subsubsection{Manual-Assisted Annotation}

This stage first adds annotations for samples with insufficient instances, and then manually corrects instances with inaccurate localization. We also remove non-realistic, spinal-fracture, and unclear samples. 

\subsubsection{Privacy Review}

Finally, we remove generated samples with excessively high comprehensive similarity (CS) to real samples in $D_{spinal2023}$. 
The values of $V_{selected}$ are assigned 
to determine whether to discard certain samples: 
\begin{equation}
v^j = \mathds{1}[\max_{r=1}^{N_r}\{CS(s^j,t^r)\} \leq \tau_{CS}], 
\end{equation}
where $CS(\cdot)$ denotes CS function, $t^r$ denotes the $r$-th sample in $D_{spinal2023}$, $N_r$ is the number of samples in $D_{spinal2023}$, and $\tau_{CS}$ is a threshold. We compare $s^j$ with all real samples so that the privacy of each real sample is preserved via $\tau_{CS}$-based selection. 
To assess the degree of patient privacy leakage in spinal image generation, we introduce the definition of memorization~\cite{carlini2023extracting}: a sample $s^j$ is considered $N_m$-completely memorized if it can be extracted from the model and there are at most $N_m$ training samples $t^r$ in the dataset such that $CS(s^j,t^r)\leq \tau_{CS}$. Here, $N_m$ denotes the number of nearly identical samples to $s^j$, which can be close to zero by setting a strict $\tau_{CS}$ value. 

To measure the similarity in terms of luminance, contrast, and structural information between two images, we use structural similarity index measure (SSIM): 
\begin{equation}  
SSIM(s^j,t^r) \!=\! \frac{(2\mu_{s^j}\mu_{t^r} + c_1)(2\sigma_{s^jt^r} + c_2)}{(\mu_{s^j}^2 + \mu_{t^r}^2 + c_1)(\sigma_{s^j}^2 + \sigma_{t^r}^2 + c_2)}, \label{eq:ssim}
\end{equation}  
where $\mu_{s^j}$ and $\sigma_{s^j}^2$ are the mean pixel value and the variance of $s^j$, respectively, $\sigma_{s^jt^r}$ is the covariance 
of $s^j$ and $t^r$, 
and $c_1$ and $c_2$ are constants to stabilize division by denominators. 
Besides, we formulate 
the pixel similarity (PS) between gray-scale transformed $s^j$ and $t^r$ with sizes $H \times W$: 
\begin{equation} 
PS(s^j,t^r) \!=\! \frac{1}{HWI} \sum_{a=1}^{H} \sum_{b=1}^{W} |GS(s^j)_{ab} - GS(t^r)_{ab}|, 
\label{eq:PS}
\end{equation} 
where $GS(\cdot)$ denotes gray-scale transformation, and $I$ denotes the number of gray scales (generally to be 255).

By combing Eqs.~\eqref{eq:ssim} and \eqref{eq:PS} using trade-off parameters $\lambda_{ss}$ and $\lambda_{ps}$, we obtain the definition of CS:
\begin{equation} 
CS(s^j,t^r)=  \lambda_{ss} SSIM(s^j,t^r)+\lambda_{ps}PS(s^j,t^r).
\label{eq:CS}
\end{equation} 
After the above selection, we can obtain currently labeled generated clean dataset $D_{spinalAI2024}$ using the selection index set $V_{selected}$.
We then merge $D_{spinalAI2024}$ with the existing $D_{spinal2023}$ to fine-tune the spinal contour detection network.
We continue to generate pseudo-labels at the next iteration, and repeat this process until $V_{selected}$ converges.

\begin{table}  
\centering  
\caption{
The statistical characteristics of different datasets.}  
\label{intro:datasets}
\resizebox{\linewidth}{!}{ 
\begin{tabular}{c|cccc|cccc|cccc}  
\toprule  
\textbf{Dataset} & \multicolumn{4}{c|}{\textbf{AASCE2019 (Public)}} & \multicolumn{4}{c|}{\textbf{Spinal2023 (Private)}} & \multicolumn{4}{c}{\textbf{Spinal-AI2024 (Generated)}} \\  
\cmidrule{2-13}  
 & \multicolumn{2}{c}{train} &\multicolumn{2}{c|}{test} & \multicolumn{2}{c}{train}  & \multicolumn{2}{c|}{test} & \multicolumn{2}{c}{train} & \multicolumn{2}{c}{test} \\  
\midrule
Number of Images& \multicolumn{2}{c}{481} & \multicolumn{2}{c|}{128} & \multicolumn{2}{c}{700} & \multicolumn{2}{c|}{204} & \multicolumn{2}{c}{16000} & \multicolumn{2}{c}{4000}  \\ 
Height of Images (Pixel)&  \multicolumn{2}{c}{973-3755} &  \multicolumn{2}{c|}{1159-2880} &  \multicolumn{2}{c}{184-833} &  \multicolumn{2}{c|}{227-802} &  \multicolumn{2}{c}{512-512} &  \multicolumn{2}{c}{512-512} \\ 

Width of Images (Pixel)&\multicolumn{2}{c}{355-1427} &  \multicolumn{2}{c|}{427-1386} &  \multicolumn{2}{c}{135-457} &  \multicolumn{2}{c|}{147-415} &  \multicolumn{2}{c}{512-512} &  \multicolumn{2}{c}{512-512} \\ 

Cobb Angle (°)&\multicolumn{2}{c}{5.96-90.0} &  \multicolumn{2}{c|}{8.21-80.99} &  \multicolumn{2}{c}{3.22-74.26} &  \multicolumn{2}{c|}{5.16-67.18} &  \multicolumn{2}{c}{0.0-90.0} &  \multicolumn{2}{c}{0.0-90.0} \\ 
\midrule
\textbf{Scoliosis Severity}&Num&Avg&Num&Avg&Num&Avg&Num&Avg&Num&Avg&Num&Avg \\ 
\midrule
Not ($<$ 10°)&17&8.35&1&8.21&91&7.82&20&8.24&1123&8.27&278&7.86\\ 
Mild (10° - 30°)&156&21.26&37&19.88&446&17.36&130&16.69&12268&18.39&2994&18.41\\ 
Moderate (30° - 45°)&156&37.53&47&38.74&76&37.87&31&37.56&1508&35.6&369&35.76\\ 
Severe ($>$ 45°)&152&60.89&43&56.98&87&55.87&23&56.1&1101&60.41&359&59.05\\ 
\bottomrule 
\end{tabular} 
}
\end{table}

\section{Experiments}

\subsection{Datasets and Settings}

\subsubsection{Datasets}

We evaluate Cobb angle estimation on public AASCE2019~\cite{wu2017automatic,wang2021evaluation} benchmark, and our private Spinal2023 and open-source Spinal-AI2024 datasets. 
The dataset details are summarized in Table \ref{intro:datasets}.
\begin{itemize}

\item \textbf{AASCE2019} contains 609 frontal spine X-ray images, in which each image is annotated by clinicians with 17 spinal segments and 68 landmarks, as well as PT, MT, and TL/L Cobb angles. 
This dataset exhibits a wide range of image intensities and Cobb angle variations. Following the official setting~\cite{wang2021evaluation}, we use 481 images for training and 128 images for testing. 

\item \textbf{Spinal2023} is our private clinical dataset, which consists of 904 spine X-ray images. 
It is randomly split into a training set of 700 images and a test set of 204 images. A public spinal landmark detection tool~\cite{yi2020vertebra} is utilized to annotate 68 landmarks of 17 spinal segments in each image, which are further manually checked and corrected by clinical experts. The PT, MT, and TL/L Cobb angles are calculated based on spinal landmarks. 
Most images have mild or moderate scoliosis, and exhibit similar image intensities.   

\item 
\textbf{Spinal-AI2024} is our generated dataset with totally 20,000 spine X-ray images, in which 16,000 and 4,000 images are used for training and testing, respectively. This dataset has a broad distribution of Cobb angles with a certain range of image intensities. The annotations of spinal landmarks as well as three Cobb angles for each image is obtained by our proposed data engine.  
\end{itemize}

\subsubsection{Implementation Details}
Our CurvNet 
is implemented via PyTorch \highlight{on an NVIDIA GeForce RTX 4090 GPU}.
Similar to 
previous work~\cite{su2024lranet}, we set the eigen-spine space dimension $M$ and the spinal segment vertex number $N$ to 16 and $14$, respectively. \highlight{The eigen-spines are constructed offline via SVD on Spinal2023 by taking 6.2 seconds, and are reused across all inference stages, ensuring good scalability.} The loss weights $\lambda_{reg}$ and $\lambda_{cls}$ in Eq.~\eqref{eq:lra_total} are set to 0.1 and 1, respectively.
We adopt stochastic gradient descent (SGD) optimizer and polynomial learning rate scheduler, with a weight decay of 0.0005, a momentum of 0.9, and an initial learning rate of 0.001.
During pseudo-labeling, our spinal contour detection network is first pre-trained for 100 epochs on Spinal2023, and then are iteratively fine-tuned on training data composed of Spinal2023 images and selected generated images. The confidence threshold $\tau_c$ is set to 0.3.
For privacy review, the weights $\lambda_{ss}$ and $\lambda_{ps}$ in Eq.~\eqref{eq:CS} are set as 0.2 and 0.8, and the CS threshold $\tau_{CS}$ is set as 0.6. \highlight{During inference, our spinal contour detection network runs at 16.7 frames per second (FPS), suitable for real-time clinical assistance.}

\highlight{For spinal image generation, we fine-tune the U-Net of Stable Diffusion v1.5 for 50 epochs using the AdamW optimizer with $\text{lr}=2\times10^{-5}$, $\beta_1=0.9$, and $\beta_2=0.999$. The used text prompts include ``AP view of human spine X-ray, scoliosis, clear vertebrae, no artifacts''. We also utilize random horizontal flips and intensity jitter for data augmentation. Latent codes are sampled from $\mathcal{N}(0, I)$, and classifier-free guidance with scale=7.5 is used to balance realism and diversity.}

\subsubsection{Evaluation Metrics}

Following the AASCE2019~\cite{wang2021evaluation} benchmark, we 
report symmetric mean absolute percentage error (SMAPE): 
\begin{equation} 
\text{SMAPE} = \frac{100\%}{N} \times \sum_{i=1}^{N} \left( \frac{|Cobb_{pr} - Cobb_{gt}|}{|Cobb_{pr}| + |Cobb_{gt}|}\right),
\end{equation} 
where $Cobb_{gt}$ and $Cobb_{pr}$ denote the ground-truth and predicted values, respectively. 
It is used to evaluate the accuracy of the maximum Cobb angle, as well as each Cobb angle of PT, MT, and TL/L.
Besides, we use object detection metrics average precision (AP) and average recall (AR), 
as well as angle difference metric Euclidean distance (ED) to assess the effect of iterative label refinement of spinal contours. 




\begin{table}[!t]  
\centering  
\caption{Comparison with state-of-the-art methods on AASCE2019 in terms of maximum Cobb angle evaluation. These methods are divided into traditional, regression-based, segmentation-based, and landmark-based methods, respectively. Our CurvNet 
can be regarded as combining regression-based and landmark-based methods.}
\label{table:comparison}
\setlength\tabcolsep{22pt}
\begin{tabular}{c| c}  

\toprule
\textbf{Method}  & \textbf{SMAPE (\%)} \\  
\midrule
S2VR \cite{sun2017direct} & 37.08 \\  
 BoostNet \cite{wu2017automatic} & 23.44 \\  
\midrule
MVC-Net \cite{wu2018automated}&35.85\\  
Faster-RCNN \cite{khanal2020automatic}& 25.7 \\  
MVE-Net \cite{wang2019accurate} & 18.95 \\  
ResNet \cite{targ2016resnet}& 10.81 \\  
\midrule
 Seg4Reg \cite{lin2020seg4reg} & 21.71 \\  
Residual U-Net \cite{horng2019cobb} & 16.48 \\  
 TOLACT \cite{chen2022automating} & 10.76 \\  
 Augmented U-Net \cite{wu2023automated} & 9.2 \\  
\midrule
CenterNet \cite{zhou2023vertebral} & 13.46 \\  
LDNet \cite{yi2020vertebra} & 10.81 \\  
HrNet \cite{sun2019deep} & 10.03 \\  
 Hourglass \cite{xu2021graph} & 9.78 \\  
Linformer \cite{guo2022cobb} & 7.91 \\  
\midrule
\textbf{CurvNet} & \textbf{5.05} \\  
\bottomrule
\end{tabular}  
\end{table}  

\subsection{Comparison with State-of-the-Art Methods}

\subsubsection{Maximum Cobb Angle Evaluation}

Table~\ref{table:comparison} shows 
comparison results on the public AASCE2019 dataset.
It can be seen that our CurvNet
maintains top-tier performance on AASCE2019. Compared to regression-based methods like Faster-RCNN, CurvNet performs better, due to our latent contour parameterized spine representation for precise Cobb angle calculation. 
Compared to segmentation-based methods, CurvNet with sparse and dense assignments can more accurately detect spinal segment contours, better addressing issues of spinal mask connectivity and fragmentation. 
Compared to landmark-based methods, our latent contour coefficient regression combined with the anchor box classification branch can more accurately localize corner points, giving stronger robustness for non-rectangular spinal segments.

\begin{figure}[!t] 
\centering
\includegraphics[width=0.6\linewidth]{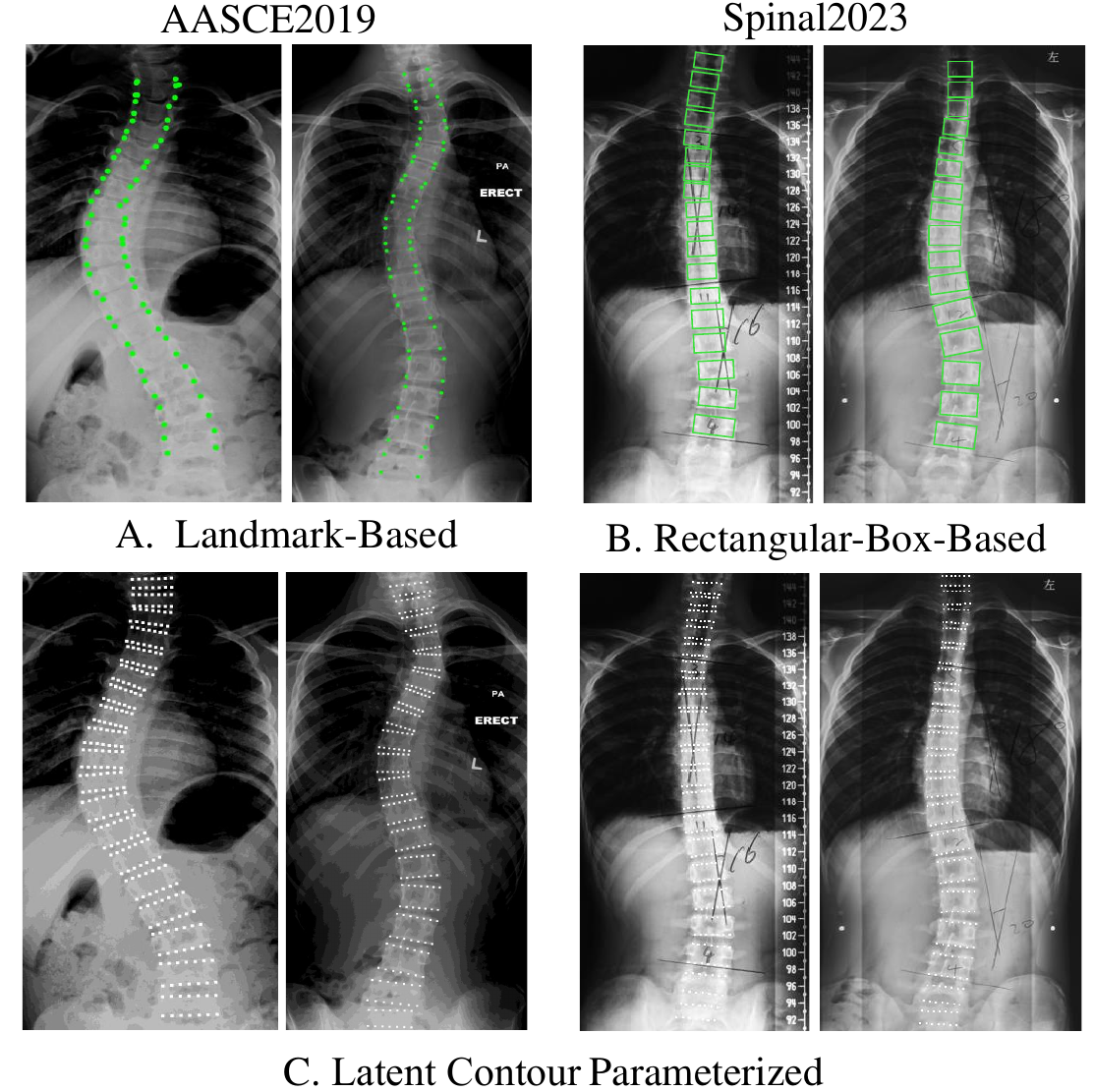}

\caption{Visualization of different spine representations on example images from AASCE2019 and Spinal2023.}
\label{page5fig5}
\end{figure}

Besides, we visualize different spine representations in Fig.~\ref{page5fig5}. It can be seen that our latent contour parameterized spine representation is more fine-grained than landmark-based and rectangular-box-based representations, enabling 
more precise Cobb angle
calculation.

\begin{table}[!t]    
\centering  
\caption{SMAPE (\%) results in terms of maximum Cobb angle and three regional Cobb angles 
on different datasets. The results on AASCE2019 are reported by literatures, and those on Spinal2023 and Spinal-AI2024 are implemented using their code.}
\resizebox{\linewidth}{!}{ 
\begin{tabular}{c|cccc|cccc|cccc}  
\toprule
\textbf{Method} & \multicolumn{4}{c|}{\textbf{AASCE2019 (Public)}} & \multicolumn{4}{c|}{\textbf{Spinal2023 (Private)}} & \multicolumn{4}{c}{\textbf{Spinal-AI2024 (Generated)}} \\  
\cmidrule{2-13}  
  & Max &PT& MT & TL/L & Max & PT &MT& TL/L & Max&PT & MT &TL/L \\  
\midrule
LDNet& 10.81 & 6.26 & 18.04 & 23.42 & 17.22 & 23.88 & 24.12 & 32.62 & 15.23 & 32.68 & 21.1 & 27.57 \\ 
Hourglass& 9.78 & 11.13 & 15.22 & 21.03 & 16.05 & 21.63 & 26.67 & 24.5 & 10.25 & 26.8 & 18.6 & 20.02 \\ 
Residual U-Net& 16.48 & 9.71 & 25.97 & 33.01 & 24.3 &  \textbf{16.6} &31.2  & 35.4 & 13.3 & 10.79 & 19.93 & 38.18 \\

Linformer& 7.91 & \textbf{4.86} & 14.65 & 20.49 & 19.1 & 17.25 & 21.17 & 29.17 & 11.59 &  \textbf{7.83} & 18.82 & 24.62 \\
\midrule 
\textbf{CurvNet} & \textbf{5.05} & 12.48 & \textbf{6.27} & \textbf{14.04} & \textbf{7.13} & 23.3 & \textbf{9.49} &  \textbf{17.25} & \textbf{5.34} & 22.91 & \textbf{8.64} & \textbf{16.69} \\  
\bottomrule
\end{tabular}  
}
\label{table:comparison-3}
\end{table}

\subsubsection{PT, MT, and TL/L Cobb Angle Evaluation}

We conduct a more detailed comparison by evaluating Cobb angles of three regional spinal curves in Table \ref{table:comparison-3}. Since the construction of Spinal-AI2024 involves our spinal contour detection network, the reported results of CurvNet on Spinal-AI2024 are obtained by re-training spinal contour detection network on Spinal-AI2024.
It can be observed that our CurvNet significantly outperforms previous works in most cases.

In particular, on 
all the three datasets, CurvNet performs the best on MT and TL/L Cobb angles, as well as maximum Cobb angle. 
This is mainly thanks to the combination of latent contour coefficient regression and anchor box
classification.
Note that the performance for PT cure is not as good, possibly because CurvNet focuses more on localizing irregular spinal segments rather than limiting the number of predicted spinal segments. As a result, it may involve the cervical spine in the calculation of PT Cobb angle, leading to worse results. Although previous methods fix the predicted whole scoliosis to 17 spinal segments, which often obtains more accurate PT predictions, complex pre-processing of data and annotations based on clinical knowledge are required.

\highlight{To more thoroughly investigate the failure causes of PT Cobb angle estimation, we analyze the failure cases by our CurvNet on the test set of AASCE2019. We define a PT estimation error as ``large'' if the absolute difference between predicted and ground-truth angle exceeds $10^\circ$. There are 37 of 128 test images with large estimation errors:
\begin{itemize}
    \item 29 cases (78\%) involve the erroneous detection of cervical vertebrae (C1–C7) as part of the proximal thoracic spine.
    \item 6 cases (16\%) are due to over-smoothing of the upper thoracic curve under poor visibility or low image contrast.
    \item 2 cases (6\%) result from severe metal artifacts near the upper spine.
\end{itemize}
78\% failure cases confirm that the main reason is the mislocalization of cervical landmarks, which extends the fitted curve cranially and leads to overestimation. To further validate this, we compute the average Cobb angle deviation when cervical landmarks are present in the predicted PT region: +14.3°, significantly larger than +3.1° when only thoracic landmarks (T1–T5) are detected.}



\highlight{Besides,} it can be seen from \highlight{Table \ref{table:comparison-3}} that segmentation-based method Residual U-Net achieves the best PT Cobb angle estimation result on Spinal2023, and landmark-based method Linformer achieves the best PT Cobb angle estimation result on Spinal-AI2024. We notice that Spinal2023 has limited images while Spinal-AI2024 has a larger scale. In this case, landmark-based methods suffer from insufficient training data on Spinal2023 but can be trained well on Spinal-AI2024, while segmentation-based methods have larger advantages on Spinal2023 with smaller scale. In contrast, our CurvNet consistently works well across small and large datasets.

\begin{table}  
\centering  
\caption{Maximum Cobb angle estimation results of different variants of our spinal contour detection network on Spinal2023. $C_{sparse}$, $C_{dense}$, $R_{sparse}$, and $R_{dense}$ denotes sparse assignment and dense assignment in the classification and regression branches, respectively. 
}
\label{table:ablation-branch}
\begin{tabular}{c c c c| c }  
\toprule
\textbf{$C_{sparse}$} & \textbf{$C_{dense}$} & \textbf{$R_{sparse}$}& \textbf{$R_{dense}$} & \textbf{SMAPE(\%)} \\  
\midrule
 \textbf{\checkmark} &  \textbf{\checkmark} &  & & 61.03  \\ 
    &  & \textbf{\checkmark}  &\textbf{\checkmark} & 53.97  \\  
 \textbf{\checkmark} &  & \textbf{\checkmark}  & & 25.02  \\ 
       & \textbf{\checkmark} &   &\textbf{\checkmark} & 17.25  \\ 
\midrule
    \textbf{\checkmark}  & \textbf{\checkmark} &  \textbf{\checkmark} &\textbf{\checkmark} & \textbf{7.13}  \\ 
\bottomrule
\end{tabular}  

\end{table}

\subsection{Ablation study}
In this section, we evaluate the key modules in our framework: spinal contour detection network and data engine. 




\subsubsection{Spinal Contour Detection Network}

Our spinal contour detection network consists of latent contour coefficient regression branch and anchor box classification branch, with each branch containing sparse and dense assignments. Table~\ref{table:ablation-branch} presents the results of different variants on Spinal2023.

\textit{Regression and Classification Branches.} We can see that using classification or regression branches alone both perform poor, in which the regression is more important. When combing both regression and classification, our full spinal contour detection network achieves the smallest SMAPE. 

\textit{Sparse and Dense Assignments.}
In the case of combining both branches, only utilizing sparse or dense assignments still cannot achieve the desired Cobb angle measurement effect. Due to more fine-grained prediction, the use of dense assignment has a higher accuracy than sparse assignment. 

\begin{table}
    \centering
    \highlight{\caption{Performance of our spinal contour detection network at different input image resolutions on AASCE2019.}
    \label{tab:resolution_performance}
    \begin{tabular}{c|cc}
    \toprule
        Resolution & SMAPE (\%) & FPS \\
        \midrule
        $512 \times 512$ & 5.05 & 16.7 \\
        Native Solutions (Averagely $ 2299.5\times 894.3$) & 5.31 & 5.6 \\
    \bottomrule
    \end{tabular}}
\end{table}

\highlight{\textit{Robustness on Input Image Resolutions.}
In our method, we resize input images to be $512 \times 512$ to feed into the spinal contour detection network. To validate the robustness of our method on resolutions of input images, we also conduct an experiment by directly feeding the native images, as presented in Table~\ref{tab:resolution_performance}. It can be observed that our spinal contour detection network still achieves comparable 5.31\% SMAPE, demonstrating good robustness. Since the average resolution of native images from AASCE2019 is $ 2299.5\times 894.3$ and is significantly larger than $512 \times 512$, the average inference speed is degraded to be 5.6 FPS. Therefore, we preserve the input $512 \times 512$ image resize setting.}    



\subsubsection{Data Engine}
During our framework, the X-ray images generated by Stable Diffusion~\cite{rombach2022high} have a certain degree of noises, in which some severe noises do not conform to reality. This problem is widespread in medical image generation, due to the distribution characteristics of training data itself, including the lack of rare cases like Cobb angles greater than 45 degrees. Our data engine is proposed to solve this issue, and is validated in this section.

\begin{figure}[!t]
\centering
\includegraphics[width=0.6\linewidth]{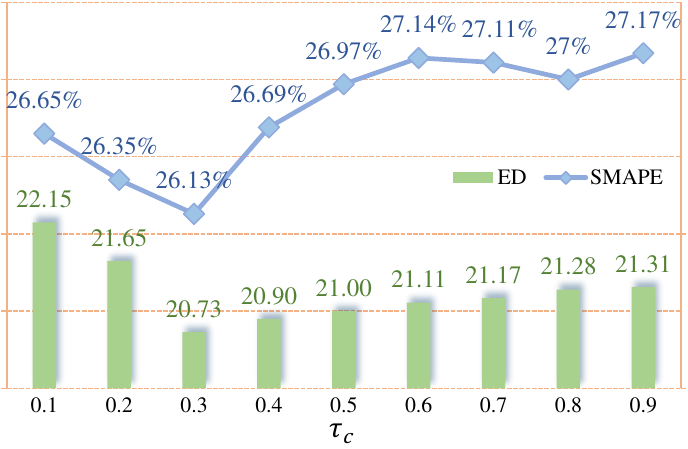}
\caption{Performance of our spinal contour detection network under different confidence thresholds on generated dataset obtained at the first data engine iteration.
Note that when $\tau_c$ is set optimally at 0.3, ED and SMAPE results on the final Spinal-AI2024 dataset will be increased from 20.73 and 26.13\% to 6.23 and 5.34\%, respectively.
}
\label{fig10}
\end{figure}

\textit{Different Confidence Thresholds.}
To investigate the effects of different confidence thresholds, we train spinal contour detection network on $D_{spinalAI2024}$ obtained at the first data engine iteration using different $\tau_c$ with a interval of 0.1, as illustrated in Fig.~\ref{fig10}.
The results show that setting $\tau_c$ to 0.3 results in the highest pseudo-label qualities. When $\tau_c$ is lower, the selected spinal segment instances may contain more noises, which complicates the further denoising process. When $\tau_c$ is higher, important positive instances may be missing, causing large errors in calculating Cobb angles. 

\begin{table}
    \centering
    \caption{\highlight{Maximum Cobb angle estimation results of our spinal contour detection network using different hyperparameters on AASCE2019.}}
\label{tab:sens_para}
    \begin{tabular}{c|c|ccc}
    \toprule
        \highlight{$\tau_{CS}=0.6$, $\lambda_{ss}/\lambda_{ps}=0.2/0.8$}&\highlight{$\tau_c$} & 0.25 & 0.30 & 0.35 \\\midrule

        \multicolumn{2}{c|}{\highlight{SMAPE (\%)}} & 5.42 & \textbf{5.05} & 5.38 \\
        \midrule\midrule
        
        \highlight{$\tau_{CS}=0.6$, $\tau_c=0.3$} &\highlight{$\lambda_{ss}/\lambda_{ps}$} & 0.1/0.9 & 0.2/0.8 & 0.5/0.5 \\\midrule

        \multicolumn{2}{c|}{\highlight{SMAPE (\%)}} & 5.12 & \textbf{5.05} & 5.18 \\
        \bottomrule

    \end{tabular}
\end{table}

\highlight{We also evaluate the sensitivity of key hyperparameters: the contour confidence threshold $\tau_c$, the consistency filtering threshold $\tau_{CS}$, and the loss weight ratio $\lambda_{ss}/\lambda_{ps}$, as shown in Table~\ref{tab:sens_para}. We set $\tau_{CS} = 0.6$ based on privacy-preserving criteria~\cite{carlini2023extracting}, where re-identification risk exceeds 80\% when $\tau_{CS} > 0.6$. It can be observed that
SMAPE remains below 6\% for $\tau_c \in [0.25, 0.35]$, with optimal performance at $\tau_c = 0.3$. For loss balancing, we use $\lambda_{ss}:\lambda_{ps} = 0.2:0.8$ to prioritize pixel similarity related to $\mathcal{L}_{ps}$, which better preserves patient-specific anatomy, and achieves the lowest SMAPE.}

\begin{table}[!t]  
\centering  
\caption{AP and AR results on six subsets of Spinal-AI2024 before and after denoising at the first data engine iteration.
}
\setlength\tabcolsep{8.5pt}
\begin{tabular}{c| c c|  c c}  
\toprule
\multirow{2}*{\textbf{Subset}}  &\multicolumn{2}{c|}{\textbf{AP}}&\multicolumn{2}{c}{\textbf{AR}}\\
\cmidrule{2-5}  
 &Before &After ($\uparrow$)&  Before& After ($\uparrow$)
  \\  
\midrule
1 &  60.5 & 79.8 \textbf{(19.3)} & 15.9 & 20.8 \textbf{(4.9)}\\ 
2 &  80.1 & 89.9 \textbf{(9.8)}  &20.9 & 26.3 \textbf{(5.4)} \\  
3 &  89.6 & 94.1 \textbf{(4.5)}&26.3 & 32.9 \textbf{(6.6)} \\ 
4 &  95.2 & 96.5 \textbf{(1.3)}&33.0 & 39.8 \textbf{(6.8)} \\ 
5 &  94.1 & 95.5 \textbf{(1.4)}&32.9 & 38.8 \textbf{(5.9)} \\ 
6 &  95.6 & 96.6 \textbf{(1.0)}&39.0 & 45.9 \textbf{(6.9)} \\ 
\bottomrule
\end{tabular}  
\label{table:ablation-denoising}
\end{table}  

\begin{table}[!t]  
\centering  
\caption{Results using different selections on Spinal-AI2024 from the first to the sixth iterations. ``No'', ``Indep.'', and ``Cumul.'' denote no filtering, and updating $V_{selected}$ in independent and cumulative ways across iterations, 
respectively.}
\begin{tabular}{c| c| c c c c c c}  
\toprule
 \multicolumn{2}{c|}{\textbf{Selection}}&1 & 2&3 & 4&5&6\\  
\midrule
 &  No &84.0	& 87.9&	95.3&	96.4&	96.6&	94.8  \\ 
 \textbf{AP}& Indep.  &\textbf{84.3}& 	88.1& 	95.5& 	96.5& 	96.6& 	96.7 \\  
  & Cumul. &\textbf{84.3}& 	\textbf{89.2}& 	\textbf{95.6}& 	\textbf{96.6}& 	\textbf{97.8}& 	\textbf{96.9} \\ 
\midrule
 & No  &26.0& 	\textbf{32.1}& 	\textbf{38.8}& 	45.5& 	56.1& 	72.8\\ 
 \textbf{AR} & Indep. &\textbf{26.1}& 	31.8& 	\textbf{38.8}&  	\textbf{45.6}&  	\textbf{56.2}& 	74.3 \\ 
  & Cumul. &\textbf{26.1}	& 32.0& 	\textbf{38.8}& 	\textbf{45.6}& 	\textbf{56.2}& 	\textbf{74.4} \\  
\bottomrule
\end{tabular}  
\label{table:ablation-selection}
\end{table}  

\textit{Denoising in Auto- and Manual-Assisted Annotation.}
Table \ref{table:ablation-denoising} shows the results before and after denoising, in which AP and AR are calculated between current labels and the labels in the final Spinal-AI2024.
It is indicated that denoising significantly improves the qualities of pseudo labels.

\begin{figure}
\centering
\includegraphics[width=0.6 \linewidth]{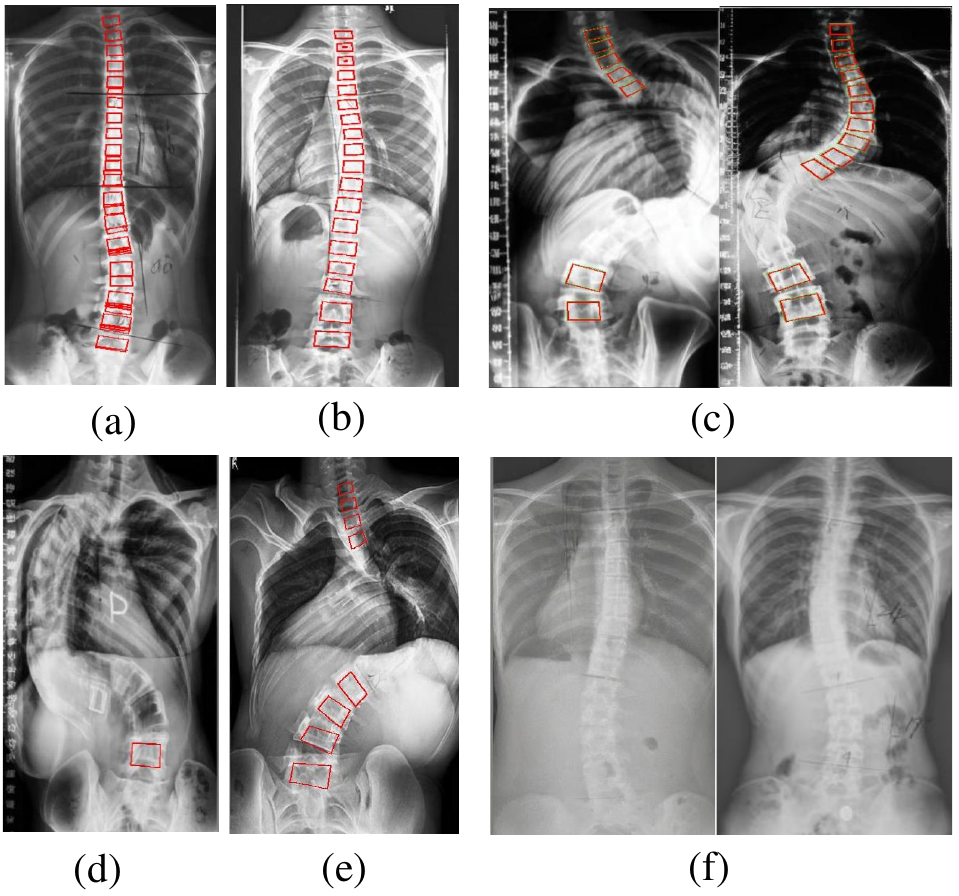} 
\caption{Illustration of discarded spinal segment instances as well as unreasonable samples during sample selection. 
(a) Inaccurate spinal segment instances due to low threshold; (b) Spinal segment instances
with area smaller than 200 pixel$^2$; (c) Samples with fewer than 10 instances; (d) Image distortion; (e) Spinal fracture; (f) Samples with
excessively high comprehensive similarity (CS). 
}
\label{page5fig7}
\end{figure}

\textit{Different Sample Selection Strategies.}
We conduct three experiments using different sample selection strategies, 
as shown in Table \ref{table:ablation-selection}. \highlight{It can be observed that the AP results of Cumul. stabilize after the fifth iteration, validating our early stopping strategy.}
If a sample is filtered and will not be selected in the next iterations, i.e. cumulative way, the overall highest quality improvement can be achieved.
If $V_{selected}$ is updated independently across iterations, the quality improvement becomes worse but is still better than no filtering. 
This demonstrates the cumulative effectiveness of combined denoising and privacy review processes.

We also present the filtered spinal segment instances and unreasonable samples during sample selection in Fig.~\ref{page5fig7}. It can be seen inaccurate pseudo-labels and poor generated samples are discarded.

\highlight{\textit{Discussion on Spinal Image Generation Model.}
In our method, we adopt latent diffusion model over GAN due to the superior mode coverage and training stability. Additionally, diffusion model offers better privacy preservation by reducing memorization risk. To validate our image generation model, we train a StyleGAN2-ADA~\cite{karras2020training} model on Spinal2023 for comparison, and compute the Fréchet inception distance (FID) score between real Spinal2023  images and generated images using a pre-trained ResNet-50. The FID score of StyleGAN2-ADA is 41.3 with severe performance dropping in extreme cases like Cobb angle larger than 45°, which is significantly worse than 28.6 FID score of our diffusion based method.}

\section{AI Responsibility and Ethical Standards}

The aim of this research is to enhance the accuracy and safety of automatic Cobb angle measurement for scoliosis diagnosis. Careful data scrutiny and adherence to ethical standards are crucial when introducing this technology. A thorough privacy review of the AI-generated data is necessary to protect patient information and rights. Upholding the highest ethical principles is a fundamental responsibility for researchers in this domain.

\subsection{Data Availability}

The datasets used in this paper include the public AASCE2019 benchmark, our private Spinal2023 dataset, and our generated Spinal-AI2024 dataset. Spinal-AI2024 has been released at \textit{https://github.com/Ernestchenchen/Spinal-AI2024}.
Although Spinal2023 can not be made publicly available due to patient privacy, this paper has provided many details regarding the data source, data processing, and data distribution. Besides, Spinal2023 is granted under a strict data usage agreement and is approved by the institutional Medical Research Ethics Committee. Spinal-AI2024 conducts privacy assessments to prevent patient re-identification.

\subsection{Data Ethics}

\subsubsection{Public AASCE2019}
The use of this benchmark dataset allows objective evaluation and comparison of our automatic Cobb angle measurement techniques. Importantly, the AASCE dataset has undergone appropriate anonymization and de-identification to protect patient privacy, in alignment with relevant ethical guidelines and regulations.

\subsubsection{Private Spinal2023}
In addition to the public AASCE dataset, this research also utilized the private Spinal2023 dataset, which contains X-ray images and Cobb angle annotations for a larger scoliosis patient cohort. 
Access to Spinal2023 was granted under a strict data usage agreement with comprehensive privacy and protection protocols, reviewed and approved by the institutional Medical Research Ethics Committee.

\subsubsection{Generated Spinal-AI2024}
This dataset was created by training generative AI models on Spinal2023 to synthesize additional X-ray images and Cobb angle annotations. The research team carefully reviewed and validated the Spinal-AI2024 dataset to ensure it preserves the original data characteristics. 

The inclusion of the Spinal-AI2024 dataset expands the training and evaluation data, which can enhance the robustness and generalizability of the automatic Cobb angle measurement algorithms. However, the use of AI-generated data requires additional ethical considerations. We have implemented rigorous data provenance tracking, algorithmic transparency, and human oversight to verify the Spinal-AI2024 dataset's integrity and fidelity. Comprehensive privacy assessments have also been conducted to prevent potential patient re-identification.

\highlight{To assess the perceptual realism of the generated Spinal-AI2024 dataset, we compute the FID score between real images from Spinal2023 and generated images from Spinal-AI2024 using a pre-trained ResNet-50, resulting in a score of 28.6, indicating good visual fidelity. Moreover, three experts independently rate 100 randomly selected generated X-ray images on a five-point Likert scale:
\begin{itemize}
    \item 1: Clearly synthetic / unrealistic
    \item 2: Mostly synthetic
    \item 3: Ambiguous
    \item 4: Mostly realistic
    \item 5: Indistinguishable from real
\end{itemize}
The distribution of scores is summarized in Table~\ref{tab:expert_eval}. The average score is 4.3, with 82\% of images scored $\ge 4$, confirming strong perceptual realism.}

\begin{table}
    \centering
    \highlight{
    \caption{Results of expert evaluation, in which three experts score 100 randomly selected generated images.}
    \label{tab:expert_eval}
    \setlength\tabcolsep{18pt}
    \begin{tabular}{cc|c}
        \toprule
        Score & Description & Percentage \\
        \midrule
        1--2 & Synthetic & 3\% \\
        3    & Ambiguous         & 15\% \\
        4    & Mostly realistic  & 52\% \\
        5    & Indistinguishable & 30\% \\
        \bottomrule
    \end{tabular}}
\end{table}

\begin{table}[!t]  \centering\caption{Image Similarity Review Checklist (excerpt). ``New Image'' denotes images from the generated dataset Spinal-AI2024, ``Top$k$-Image'' denotes the image in the $k$-th place of descending similarity 
in the private dataset Spinal2023, and ``ACS'' denotes the average of Top1-CS, Top2-CS, and Top3-CS. 
}  
\label{tab:SRC}
\resizebox{\linewidth}{!}{ 
\begin{tabular}{c|ccc|ccc|ccc|c}  \toprule  
New Image&Top1-Image&Top1-SSIM& $\cdots$ &Top1-Image&Top1-PS&$\cdots$&Top1-CS&Top2-CS&Top3-CS&\textbf{ACS} \\  
\midrule  
  000001.jpg & 000196.png&38.24 &$\cdots$&000290.png& 37.78 & $\cdots$ & 47.61 & 47.15 & 46.77&47.18 \\  
\vdots & \vdots & \vdots & \vdots & \vdots & \vdots & \vdots & \vdots & \vdots & \vdots & \vdots \\  
001257.jpg&000527.png&\textbf{52.76}&$\cdots$&000427.png&88.58&$\cdots$&\textbf{59.92}&58.18&58.01&\textbf{58.71} \\  
\vdots & \vdots & \vdots & \vdots & \vdots & \vdots & \vdots & \vdots & \vdots & \vdots & \vdots \\ 
020140.jpg&000009.png&44.2&$\cdots$&000531.png&\textbf{89.3}&$\cdots$&53.22&53.15&52.83&53.07\\  
\vdots & \vdots & \vdots & \vdots & \vdots & \vdots & \vdots & \vdots & \vdots & \vdots & \vdots \\ 
 020000.jpg &000763.png&34.46&$\cdots$&000493.png&82.25&$\cdots$&44.02&43.91&43.88&43.93 \\ 
\bottomrule  
\end{tabular} 
}
\end{table} 


\begin{figure}[!t]
\centering
\includegraphics[width=0.6\linewidth]{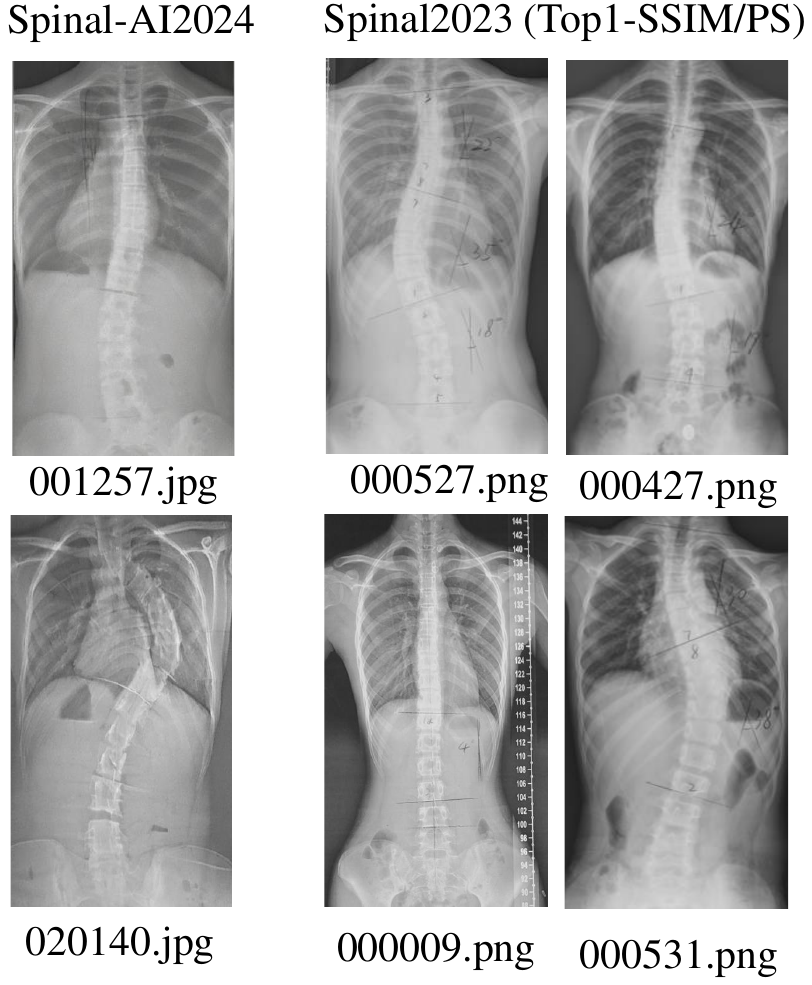}
\caption{Illustration of highest similar images.}
\label{fig:highest-sim}
\end{figure}

\highlight{Besides,} for each Spinal-AI2024 image, a top-3 comprehensive similarity review is presented in Table~\ref{tab:SRC}, 
in which the most similar images are shown in Fig.~\ref{fig:highest-sim}. The results show that our generated dataset has distorted the original annotations and produced spinal curvature variations, which can prevent excessive patient information leakage.

\section{Conclusion}
We have proposed a novel curvature angle estimation framework with latent contour representation based contour detection and iterative data engine based image self-generation.
Latent contour coefficient regression and anchor box classification with sparse and dense assignments are combined to detect irregular contours of spinal segments.
Besides, we have proposed an iterative data engine with semi-supervised labeling and privacy review, which culminates in the open-source largest scoliosis X-ray dataset Spinal-AI2024. Extensive experiments have demonstrated the effectiveness of our method, achieving state-of-the-art performance on public, private, and generated datasets for both maximum Cobb angle and three regional Cobb angles. \highlight{In future work, we will improve PT Cobb angle estimation by introducing anatomical constraints or region-specific landmark filtering.}

\section*{CRediT authorship contribution statement}
\textbf{Zhiwen~Shao:} Writing – review \& editing, Methodology, Supervision, Funding acquisition, Conceptualization.
\textbf{Yichen~Yuan:} Writing - original draft, Visualization, Investigation, Software.
\textbf{Lizhuang~Ma:} Formal analysis, Validation, Funding acquisition.
\textbf{Xiaojia~Zhu:} Writing – review \& editing, Data curation, Resources, Project administration.

\section*{Declaration of competing interest}

The authors declare that they have no known competing financial interests or personal relationships that could have appeared to influence the work reported in this paper.

\section*{Acknowledgments}
This work was supported in part by the National Natural Science Foundation of China under Grants 62472424, 62472282, and 72192821, in part by the China Postdoctoral Science Foundation under Grant 2023M732223, and in part by the Hong Kong Scholars Program under Grant XJ2023037/HKSP23EG01.

\section*{Data availability}

The used AASCE2019 and Spinal-AI2024
can be downloaded at \textit{https://aasce19.grand-challenge.org} and \textit{https://github.com/Ernestchenchen/Spinal-AI2024}, respectively. The used Spinal2023 is private due to privacy protection. Our code can be downloaded at \textit{https://github.com/Ernestchenchen/CurvNet}.

 \bibliographystyle{elsarticle-num} 
 \bibliography{reference.bib}

\begin{thebibliography}{10}
\expandafter\ifx\csname url\endcsname\relax
  \def\url#1{\texttt{#1}}\fi
\expandafter\ifx\csname urlprefix\endcsname\relax\def\urlprefix{URL }\fi
\expandafter\ifx\csname href\endcsname\relax
  \def\href#1#2{#2} \def\path#1{#1}\fi

\bibitem{zhu2025mgscoliosis}
X.~Zhu, R.~Chen, Z.~Shao, M.~Zhang, Y.~Dai, W.~Zhang, C.~Lang, Mgscoliosis: Multi-grained scoliosis detection with joint ordinal regression from natural image, Alexandria Engineering Journal 111 (2025) 329--340.

\bibitem{prabhu2012automatic}
G.~Prabhu, Automatic quantification of spinal curvature in scoliotic radiograph using image processing, Journal of Medical Systems 36~(3) (2012) 1943--1951.

\bibitem{mcleod2022anesthesia}
C.~B. McLeod, Anesthesia for pediatric spinal deformity, in: Multidisciplinary Spine Care, Springer, 2022, pp. 667--710.

\bibitem{sun2022comparison}
Y.~Sun, Y.~Xing, Z.~Zhao, X.~Meng, G.~Xu, Y.~Hai, Comparison of manual versus automated measurement of cobb angle in idiopathic scoliosis based on a deep learning keypoint detection technology, European Spine Journal~(31) (2022) 1969--1978.

\bibitem{wu2017automatic}
H.~Wu, C.~Bailey, P.~Rasoulinejad, S.~Li, Automatic landmark estimation for adolescent idiopathic scoliosis assessment using boostnet, in: International Conference on Medical Image Computing and Computer Assisted Intervention, Springer, 2017, pp. 127--135.

\bibitem{khanal2020automatic}
B.~Khanal, L.~Dahal, P.~Adhikari, B.~Khanal, Automatic cobb angle detection using vertebra detector and vertebra corners regression, in: Computational Methods and Clinical Applications for Spine Imaging, Springer, 2020, pp. 81--87.

\bibitem{lin2020seg4reg}
Y.~Lin, H.-Y. Zhou, K.~Ma, X.~Yang, Y.~Zheng, Seg4reg networks for automated spinal curvature estimation, in: Computational Methods and Clinical Applications for Spine Imaging, Springer, 2020, pp. 69--74.

\bibitem{wu2023automated}
Y.~Wu, K.~Namdar, C.~Chen, S.~Hosseinpour, M.~Shroff, A.~S. Doria, F.~Khalvati, Automated adolescence scoliosis detection using augmented u-net with non-square kernels, Canadian Association of Radiologists Journal 74~(4) (2023) 667--675.

\bibitem{xu2021graph}
T.~Xu, W.~Takano, Graph stacked hourglass networks for 3d human pose estimation, in: IEEE Conference on Computer Vision and Pattern Recognition, IEEE, 2021, pp. 16105--16114.

\bibitem{yi2020vertebra}
J.~Yi, P.~Wu, Q.~Huang, H.~Qu, D.~N. Metaxas, Vertebra-focused landmark detection for scoliosis assessment, in: IEEE International Symposium on Biomedical Imaging, IEEE, 2020, pp. 736--740.

\bibitem{su2024lranet}
Y.~Su, Z.~Chen, Z.~Shao, Y.~Du, Z.~Ji, J.~Bai, Y.~Zhou, Y.-G. Jiang, Lranet: Towards accurate and efficient scene text detection with low-rank approximation network, in: AAAI Conference on Artificial Intelligence, 2024, pp. 4979--4987.

\bibitem{rombach2022high}
R.~Rombach, A.~Blattmann, D.~Lorenz, P.~Esser, B.~Ommer, High-resolution image synthesis with latent diffusion models, in: IEEE Conference on Computer Vision and Pattern Recognition, IEEE, 2022, pp. 10684--10695.

\bibitem{carlini2023extracting}
N.~Carlini, J.~Hayes, M.~Nasr, M.~Jagielski, V.~Sehwag, F.~Tramer, B.~Balle, D.~Ippolito, E.~Wallace, Extracting training data from diffusion models, in: USENIX Security Symposium, 2023, pp. 5253--5270.

\bibitem{packhauser2023generation}
K.~Packh{\"a}user, L.~Folle, F.~Thamm, A.~Maier, Generation of anonymous chest radiographs using latent diffusion models for training thoracic abnormality classification systems, in: IEEE International Symposium on Biomedical Imaging (ISBI), IEEE, 2023, pp. 1--5.

\bibitem{rizve2021defense}
M.~N. Rizve, K.~Duarte, Y.~S. Rawat, M.~Shah, In defense of pseudo-labeling: An uncertainty-aware pseudo-label selection framework for semi-supervised learning, International Conference on Learning Representations (2021).

\bibitem{kirillov2023segment}
A.~Kirillov, E.~Mintun, N.~Ravi, H.~Mao, C.~Rolland, L.~Gustafson, T.~Xiao, S.~Whitehead, A.~C. Berg, W.-Y. Lo, et~al., Segment anything, in: IEEE International Conference on Computer Vision, IEEE, 2023, pp. 4015--4026.

\bibitem{sun2017direct}
H.~Sun, X.~Zhen, C.~Bailey, P.~Rasoulinejad, Y.~Yin, S.~Li, Direct estimation of spinal cobb angles by structured multi-output regression, in: International Conference on Information Processing in Medical Imaging, Springer, 2017, pp. 529--540.

\bibitem{shao2025mol}
Z.~Shao, Y.~Cheng, F.~Li, Y.~Zhou, X.~Lu, Y.~Xie, L.~Ma, Mol: Joint estimation of micro-expression, optical flow, and landmark via transformer-graph-style convolution, IEEE Transactions on Pattern Analysis and Machine Intelligence (2025).

\bibitem{targ2016resnet}
S.~Targ, D.~Almeida, K.~Lyman, Resnet in resnet: Generalizing residual architectures, in: International Conference on Learning Representations Workshops, 2016.

\bibitem{wu2018automated}
H.~Wu, C.~Bailey, P.~Rasoulinejad, S.~Li, Automated comprehensive adolescent idiopathic scoliosis assessment using mvc-net, Medical Image Analysis 48 (2018) 1--11.

\bibitem{wang2019accurate}
L.~Wang, Q.~Xu, S.~Leung, J.~Chung, B.~Chen, S.~Li, Accurate automated cobb angles estimation using multi-view extrapolation net, Medical Image Analysis 58 (2019) 101542.

\bibitem{chen2022automating}
C.~Chen, K.~Namdar, Y.~Wu, S.~Hosseinpour, M.~Shroff, A.~Doria, F.~Khalvati, Automating cobb angle measurement for adolescent idiopathic scoliosis using instance segmentation, in: Annual International Conference of the IEEE Engineering in Medicine and Biology Society, IEEE, 2024, pp. 1--5.

\bibitem{ronneberger2015u}
O.~Ronneberger, P.~Fischer, T.~Brox, U-net: Convolutional networks for biomedical image segmentation, in: International Conference on Medical Image Computing and Computer Assisted Intervention, Springer, 2015, pp. 234--241.

\bibitem{horng2019cobb}
M.-H. Horng, C.-P. Kuok, M.-J. Fu, C.-J. Lin, Y.-N. Sun, Cobb angle measurement of spine from x-ray images using convolutional neural network, Computational and Mathematical Methods in Medicine 2019~(1) (2019) 6357171.

\bibitem{sun2019deep}
K.~Sun, B.~Xiao, D.~Liu, J.~Wang, Deep high-resolution representation learning for human pose estimation, in: IEEE Conference on Computer Vision and Pattern Recognition, IEEE, 2019, pp. 5693--5703.

\bibitem{zhou2023vertebral}
Z.~Zhou, J.~Zhu, C.~Yao, Vertebral center points locating and cobb angle measurement based on deep learning, Applied Sciences 13~(6) (2023) 3817.

\bibitem{guo2022cobb}
Y.~Guo, Y.~Li, H.~Song, W.~He, K.~Yuan, Cobb angle rectification with dual-activated linformer, in: IEEE International Conference on Mechatronics and Automation, IEEE, 2022, pp. 1003--1008.

\bibitem{goodfellow2020generative}
I.~Goodfellow, J.~Pouget-Abadie, M.~Mirza, B.~Xu, D.~Warde-Farley, S.~Ozair, A.~Courville, Y.~Bengio, Generative adversarial networks, Communications of the ACM 63~(11) (2020) 139--144.

\bibitem{esteva2019guide}
A.~Esteva, A.~Robicquet, B.~Ramsundar, V.~Kuleshov, M.~DePristo, K.~Chou, C.~Cui, G.~Corrado, S.~Thrun, J.~Dean, A guide to deep learning in healthcare, Nature Medicine 25~(1) (2019) 24--29.

\bibitem{honari2018improving}
S.~Honari, P.~Molchanov, S.~Tyree, P.~Vincent, C.~Pal, J.~Kautz, Improving landmark localization with semi-supervised learning, in: IEEE Conference on Computer Vision and Pattern Recognition, IEEE, 2018, pp. 1546--1555.

\bibitem{li2021synthetic}
C.~Li, G.~H. Lee, From synthetic to real: Unsupervised domain adaptation for animal pose estimation, in: IEEE Conference on Computer Vision and Pattern Recognition, IEEE, 2021, pp. 1482--1491.

\bibitem{shi2018transductive}
W.~Shi, Y.~Gong, C.~Ding, Z.~M. Tao, N.~Zheng, Transductive semi-supervised deep learning using min-max features, in: European Conference on Computer Vision, 2018, pp. 299--315.

\bibitem{yi2019probabilistic}
K.~Yi, J.~Wu, Probabilistic end-to-end noise correction for learning with noisy labels, in: IEEE Conference on Computer Vision and Pattern Recognition, IEEE, 2019, pp. 7017--7025.

\bibitem{wang2020repetitive}
G.-H. Wang, J.~Wu, Repetitive reprediction deep decipher for semi-supervised learning, in: AAAI Conference on Artificial Intelligence, 2020, pp. 6170--6177.

\bibitem{xie2020self}
Q.~Xie, M.-T. Luong, E.~Hovy, Q.~V. Le, Self-training with noisy student improves imagenet classification, in: IEEE Conference on Computer Vision and Pattern Recognition, IEEE, 2020, pp. 10687--10698.

\bibitem{he2016deep}
K.~He, X.~Zhang, S.~Ren, J.~Sun, Deep residual learning for image recognition, in: IEEE Conference on Computer Vision and Pattern Recognition, IEEE, 2016, pp. 770--778.

\bibitem{lin2017feature}
T.-Y. Lin, P.~Doll{\'a}r, R.~Girshick, K.~He, B.~Hariharan, S.~Belongie, Feature pyramid networks for object detection, in: IEEE Conference on Computer Vision and Pattern Recognition, IEEE, 2017, pp. 2117--2125.

\bibitem{blum2020foundations}
A.~Blum, J.~Hopcroft, R.~Kannan, Foundations of Data Science, Cambridge University Press, 2020.

\bibitem{wang2021evaluation}
L.~Wang, C.~Xie, Y.~Lin, H.-Y. Zhou, K.~Chen, D.~Cheng, F.~Dubost, B.~Collery, B.~Khanal, B.~Khanal, et~al., Evaluation and comparison of accurate automated spinal curvature estimation algorithms with spinal anterior-posterior x-ray images: The aasce2019 challenge, Medical Image Analysis 72 (2021) 102115.

\bibitem{karras2020training}
T.~Karras, M.~Aittala, J.~Hellsten, S.~Laine, J.~Lehtinen, T.~Aila, Training generative adversarial networks with limited data, in: Advances in Neural Information Processing Systems, 2020, pp. 12104--12114.

\end{thebibliography}

\end{document}